\documentclass[11pt]{article}

\usepackage[final]{acl}

\usepackage{times}
\usepackage{latexsym}

\usepackage[LAE, T1]{fontenc}
\usepackage[utf8]{inputenc}
\usepackage[bidi=default]{babel}
\usepackage{microtype}
\usepackage{amsmath}
\usepackage{inconsolata}
\usepackage{booktabs}
\usepackage{float}
\usepackage{graphicx}
\usepackage{enumitem}
\usepackage{listings}
\usepackage{xcolor}
\usepackage{hyperref}

\title{REGLAT at ArGuard Shared Tasks: An Ensemble-Based Model Integrating NER for Harmful Content Detection in Arabic}

\author{Ahmed M. Fetouh\textsuperscript{1}\and Mariam Labib\textsuperscript{2}\and Nsrin Ashraf\textsuperscript{1,2}\and Omer Dawood\textsuperscript{3}\\ \textbf{and Hamada Nayel}\textsuperscript{1,3}\\
\textsuperscript{1}Faculty of Computers and Artificial Intelligence, Benha University, Egypt \\
\textsuperscript{2}Computer Engineering, Elsewedy University of Technology, Cairo, Egypt \\
\textsuperscript{3}College of Engineering, Wadi Ad Dwaser,\\ Prince Sattam Bin Abdulaziz University, KSA}
\begin{document}
\maketitle
\begin{abstract}
Detecting harmful content in Arabic is challenging because of dialectal variations and the semantic overlap between different harm categories. This paper presents a two-stage approach for classifying Arabic user prompts into seven categories of harmful content. First, GLiNER is used to extract harm-related entities from each prompt. The extracted entities are then added to the original text and passed to a fine-tuned MARBERT classifier for final classification. The approach combines explicit entity information with the contextual understanding of a pretrained Arabic language model. The system was evaluated on the development set of the ArGuard 2026 shared task. The results achieved a macro-F1 score of 0.886 and an accuracy of 0.897, demonstrating promising performance in fine-grained harmful content detection. The system reported an accuracy of 0.793 and Macro-F score of 0.735 on a blind test set for subtask 1. For subtask 2, it achieved an accuracy of 0.770 and Macro-F score of 0.754. The analysis also showed that distinguishing between related categories remains challenging. These findings suggest that integrating NER with Arabic language models is a promising approach to harmful content detection.
\end{abstract}
%--------------------------------------------------------------
\section{Introduction}
The rapid adoption of large language models (LLMs) and conversational artificial intelligence has increased the importance of detecting harmful content in user prompts. Although these systems provide valuable capabilities for information retrieval, content generation, and interactive assistance, they may also encounter prompts involving violence, self-harm, harassment, bullying, hate speech, fraud, and adult content. Reliable harmful content detection is therefore essential for improving the safety and responsible use of language technologies. However, identifying harmful content remains challenging, particularly in Arabic, where dialectal variation, informal expressions, and context-dependent meanings can complicate automated classification~\cite{ashraf-etal-2025-reglat,9781707,nayel-etal-2021-machine-learning}.

Fine-grained harmful content detection presents additional challenges beyond distinguishing between safe and unsafe text. Systems must identify the specific category of harmful content expressed in a prompt, even when different categories share similar vocabulary or semantic characteristics. For example, bullying, harassment, and hate speech may all involve offensive language or threats, while self-harm and harm to others may share references to violence or physical injury. Furthermore, a prompt may contain potentially harmful entities without explicitly expressing harmful intent, making it necessary to consider both the words used and their surrounding context. Class imbalance can further complicate this task, as models may perform unevenly across categories with different numbers of training examples~\cite{nayel-2020-nayel}.

Transformer-based language models have demonstrated their usefulness in text classification by learning contextual representations from large-scale pretraining. MARBERT is relevant to Arabic text because its pretraining includes Modern Standard Arabic and dialectal Arabic data~\cite{abdul-mageed-etal-2021-arbert}. Nevertheless, contextual representations may not always capture sufficiently explicit entity-level information needed to distinguish among harmful content categories. Named entity recognition (NER) offers a complementary approach by identifying relevant entities and expressions, such as weapons, self-injury references, sensitive financial information, and offensive expressions~\cite{ELDOUBY2026618}. Integrating these extracted entities into a classification pipeline may provide additional structured cues to support fine-grained decisions.

%We propose a two-stage system for ArGuard 2026 Subtask B2 that combines GLiNER entity extraction with a fine-tuned MARBERT classifier. Extracted harm-related entities are encoded as structured text and appended to the original prompt before classification into seven harmful-content categories. Results demonstrate the value of integrating explicit entity information with contextual representations, although distinguishing closely related categories such as harm to others, bullying, and harassment remains challenging.
This paper presents our system for Subtask B2 of ArGuard 2026~\cite{alam-etal-2026-arguard}, which combines GLiNER-based entity extraction with a fine-tuned MARBERT classifier. Harm-related entities are first extracted and transformed into a structured representation that is appended to the original prompt. The enriched input is then classified by MARBERT into one of seven harmful-content categories. By integrating explicit entity information with contextual language representations, the approach improves category discrimination. While achieving strong results on categories such as adult content and hate speech, error analysis reveals persistent confusion among semantically similar classes, particularly harm to others, bullying, and harassment, highlighting the ongoing challenge of fine-grained harmful content detection in Arabic.
The main contributions of this work are as follows:-
\begin{itemize}
    \item It presents a two-stage framework that integrates generalist NER with Arabic transformer-based classification for fine-grained harmful content detection,
    \item it introduces a structured entity-augmented representation that makes harm-related entity cues explicit within the classifier's input,
    \item it provides an empirical evaluation and error analysis on the ArGuard 2026 development set, identifying category-specific strengths and the principal sources of classification errors.
\end{itemize}

The remainder of this paper is organized as follows. Section 2 describes the shared-task objectives and dataset. Section 3 presents the overall system architecture, followed by the entity extraction and input representation methods. The MARBERT classification approach and experimental configuration are then described. Finally, the paper reports the evaluation results, discusses classification errors and ablation experiments, and concludes with directions for future work.
\section{Task and Data}
\label{sec:data}
\begin{figure}[!t]
\centering
\includegraphics[width=\linewidth]{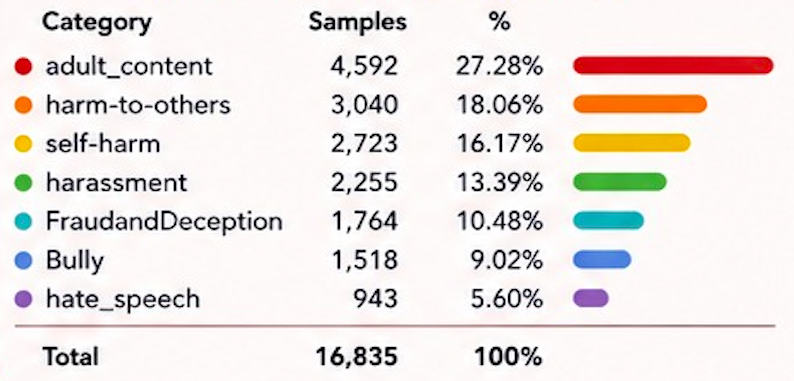}
\caption{Granular category distribution in the training set.}
\label{fig:statistics}
\end{figure}
Subtask B2 of ArGuard 2026~\cite{alam-etal-2026-arguard}, asks systems to assign each Arabic user prompt to one of seven granular harm categories: \texttt{adult\_content}, \texttt{harm-to-others}, \texttt{self-harm}, \texttt{harassment}, \texttt{FraudandDeception}, \texttt{Bully}, and \texttt{hate\_speech}. The training set contains 16,835 prompts~\cite{kmainasi2026ahamemes}. Each sample consists of a prompt, a binary safety label (safe/unsafe), a granular label, and a dialect tag (e.g., Modern Standard Arabic and Levantine). The class distribution is imbalanced: \texttt{adult\_content} is the largest class (4,592 samples, 27.28\%), while \texttt{hate\_speech} is the smallest (943 samples, 5.60\%), an imbalance ratio of roughly 4.9:1. The task is difficult because several categories overlap semantically (for example, harassment, bullying, and hate speech), because harmful intent is often expressed through dialectal or idiomatic wording, and because surface-benign prompts can still belong to a sensitive category.

Before modeling, we inspected missing values, duplicate prompts and identifiers, and prompt length statistics (in characters and words). Rows with missing values were removed from the training set. The labelled development set was used for model selection and evaluation, and the unlabeled test set was used only for final prediction.

%--------------------------------------------------------------
\section{System Overview}
\label{sec:overview}

Our system is a two-stage pipeline that couples NER with a pretrained Arabic transformer. In the first stage, GLiNER~\cite{zaratiana-etal-2024-gliner} extracts harm-related entities from the prompt. In the second stage, the extracted entities are serialized as text and appended to the original prompt, and the enriched text is classified by a fine-tuned MARBERT model. The intuition is that explicit entity cues (weapons, self-injury terms, financial credentials, protected groups, insults) give the classifier a compact, category-oriented summary of the prompt that complements the contextual representation learned by the encoder. The same pipeline is applied unchanged to the training, development, and test sets.
\begin{figure*}[!t]
\centering
\includegraphics[width=\textwidth]{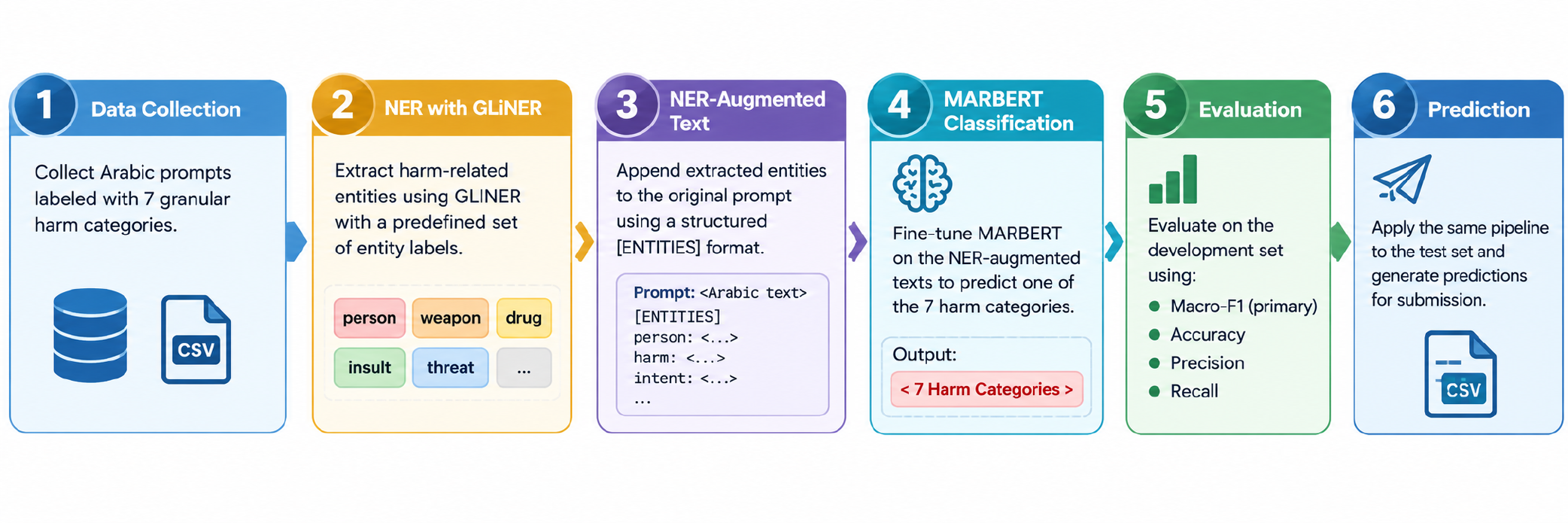}
\caption{The pipeline of the proposed system.}
\label{fig:pipeline}
\end{figure*}

%--------------------------------------------------------------
\subsection{Entity Extraction with GLiNER}
\label{sec:gliner}
We use GLiNER, a generalist NER model that accepts an arbitrary label set at inference time, with the multilingual checkpoint\footnote{\texttt{urchade/gliner\_multi-v2.1}}. Because GLiNER is zero-shot with respect to the label set, no NER-specific training data is required. We define 34 entity labels organized into seven harm domains aligned with the task categories:
\begin{itemize}
  \item \textbf{Violence and physical harm:} weapon, firearm, explosive, bomb, knife, poison, violent action, physical harm.
  \item \textbf{Self-harm:} suicide, suicide method, self harm, self injury.
  \item \textbf{Fraud and sensitive information:} credit card, bank account, financial information, password, credential, identity document, personal information.
  \item \textbf{Adult content:} sexual activity, pornographic content, nudity, explicit content.
  \item \textbf{Hate:} ethnic group, religious group, nationality, protected group.
  \item \textbf{Harassment:} insult, threat, bullying, offensive expression.
  \item \textbf{Contextual entities:} person, organization, location.
\end{itemize}
Entities are predicted with a confidence threshold of 0.5. For every detected entity we store its surface text, confidence score, and character offsets. Duplicate entities within the same label are collapsed by surface form, yielding a unique set of entities per label for each prompt. The results are saved in JSONL format together with the prompt identifier and, where available, the gold labels.

%--------------------------------------------------------------
\subsection{NER-Augmented Text Representation}
\label{sec:augmented}
The structured entity output is converted into a plain-text block that follows the original prompt. For every one of the 34 labels, in a fixed order, we write one line of the form \texttt{label: entity\_1, entity\_2, ...}, or \texttt{label: none} if nothing was detected. The final model input is:

\begin{lstlisting}
<original prompt>
[ENTITIES]
weapon: <entities or none>
firearm: <entities or none>
...
location: <entities or none>
\end{lstlisting}

The fixed ordering and explicit \texttt{none} values keep the input format identical across samples, so that the presence or absence of a cue in a given domain is directly visible to the classifier. Granular labels are mapped to integer identifiers with a label encoder fitted on the training set; the same encoder is reused for the development set and to decode test predictions.

%--------------------------------------------------------------
\subsection{MARBERT Classifier}
\label{sec:marbert}

We fine-tune MARBERT (\texttt{UBC-NLP/MARBERT}), a BERT-based model pretrained on large-scale Modern Standard Arabic and dialectal Arabic text, which suits the dialectal variation present in the data. We attach a sequence-classification head with seven output units and train with the standard cross-entropy objective, without class weighting or resampling. Inputs are tokenized with the MARBERT tokenizer, truncated and padded to a maximum length of 256 tokens. Only the granular label is predicted; the binary safe/unsafe label is not used as an input or auxiliary target.

\subsection{Training Configuration}

Training uses the \texttt{Hugging Face Trainer}\footnote{https://huggingface.co/docs/transformers/en/trainer} with the hyperparameters in Table~\ref{tab:hparams}. After each epoch the model is evaluated on the development set, and the checkpoint with the highest macro-F1 is restored at the end of training.

\begin{table}[t]
\centering
\caption{Training hyperparameters.}
\label{tab:hparams}
\begin{tabular}{ll}\hline
Hyperparameter & Value \\ \hline
Learning rate                   & $2\times10^{-5}$ \\
Warmup          & $10\%$  \\
Epochs                          & 3 \\
Train / eval batch size         & 32 / 16 \\
Weight decay                    & 0.01 \\
Model selection                 & Macro-F1 \\
\hline
\end{tabular}
\end{table}

%--------------------------------------------------------------
\section{Evaluation and Inference}
\label{sec:eval}

The official primary metric is macro-F1, which weights all seven categories equally and is therefore appropriate under class imbalance. We additionally report accuracy, macro-averaged precision, and macro-averaged recall, together with per-class precision, recall, and F1 and raw and row-normalized confusion matrices to analyze inter-class confusion.

At test time, each prompt passes through the identical pipeline: GLiNER entity extraction with the same label set and threshold, construction of the \texttt{[ENTITIES]}-augmented text, tokenization with the MARBERT tokenizer, and prediction with the selected checkpoint. The predicted class identifiers are mapped back to granular label names with the training label encoder.

%--------------------------------------------------------------
\section{Results}
\label{sec:results}

Table~\ref{tab:results} summarizes the development-set performance of the GLiNER + MARBERT system on the 2,405 development prompts. The system reaches a macro-F1 of 0.886 and an accuracy of 0.897. Macro-averaged precision and recall are close (0.885 and 0.888), indicating that the model is not systematically biased toward over- or under-predicting any category.

\begin{table}[t]
\centering
\caption{Overall development-set results.}
\label{tab:results}
\begin{tabular}{lc}
\toprule
Metric & Score \\
\midrule
Macro-F1        & 0.886 \\
Accuracy        & 0.897 \\
Macro-Precision & 0.885 \\
Macro-Recall    & 0.888 \\
\bottomrule
\end{tabular}
\end{table}

Table~\ref{tab:perclass} reports per-class results. \texttt{adult\_content} is almost perfectly recognized (F1 = 0.987), and \texttt{hate\_speech} also performs strongly (F1 = 0.923) despite being the smallest training class, which suggests that the entity cues for protected groups and offensive expressions are informative. The weakest class is \texttt{harm-to-others} (F1 = 0.815), followed by \texttt{Bully} (0.834) and \texttt{harassment} (0.848).

\begin{table*}[t]
\centering
\caption{Per-class development-set classification report of the GLiNER + MARBERT system.}
\label{tab:perclass}
\begin{tabular}{lcccr}\hline
Category & Precision & Recall & F1 & Support \\\hline
\texttt{Bully}              & 0.8341 & 0.8341 & 0.8341 & 217 \\
\texttt{FraudandDeception}  & 0.8842 & 0.9087 & 0.8963 & 252 \\
\texttt{adult\_content}     & 0.9863 & 0.9878 & 0.9871 & 657 \\
\texttt{harassment}         & 0.8457 & 0.8509 & 0.8483 & 322 \\
\texttt{harm-to-others}     & 0.8231 & 0.8060 & 0.8145 & 433 \\
\texttt{hate\_speech}       & 0.9191 & 0.9259 & 0.9225 & 135 \\
\texttt{self-harm}          & 0.9044 & 0.8997 & 0.9021 & 389 \\\hline
Accuracy                    & \multicolumn{3}{c}{0.8969} & 2405 \\
Macro average               & 0.8853 & 0.8876 & 0.8864 & 2405 \\
Weighted average            & 0.8966 & 0.8969 & 0.8967 & 2405 \\\hline
\end{tabular}
\end{table*}

\subsection{Error Analysis}
\begin{figure}[t]
\centering
\includegraphics[width=\linewidth]{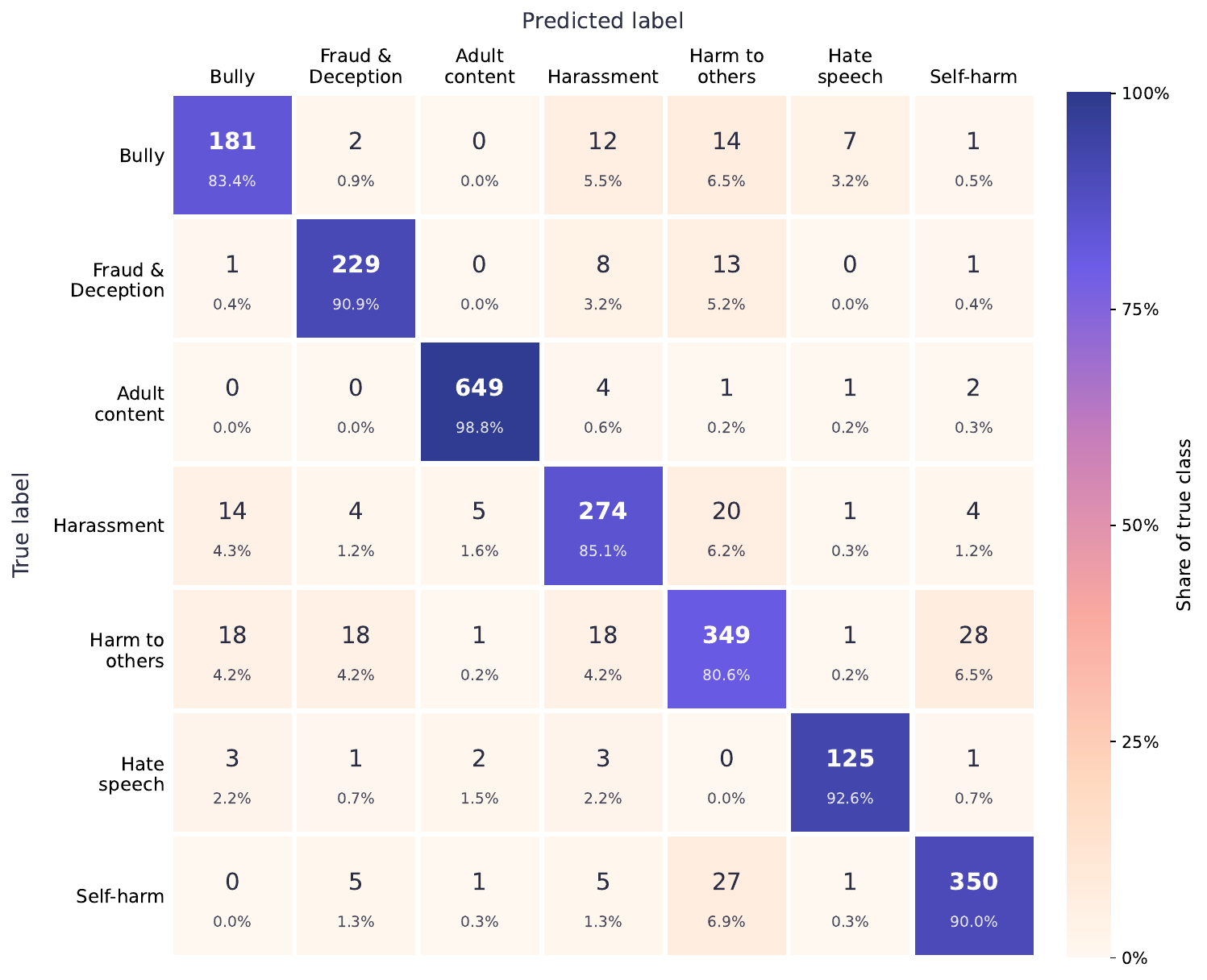}
\caption{Development-set confusion matrix of the GLiNER + MARBERT system.}
\label{fig:cm}
\end{figure}

As shown in Figure~\ref{fig:cm}, errors concentrate among semantically adjacent categories rather than being spread uniformly. The largest confusion is between \texttt{harm-to-others} and \texttt{self-harm}: 28 \texttt{harm-to-others} prompts are predicted as \texttt{self-harm} and 27 \texttt{self-harm} prompts as \texttt{harm-to-others}, which is plausible since both involve physical harm, weapons, and violent actions, and differ mainly in who the target is. \texttt{harm-to-others} is also confused with \texttt{Bully}, \texttt{FraudandDeception}, and \texttt{harassment} (18 prompts each), so it acts as a hub class that absorbs errors from several neighbors. A second cluster involves \texttt{Bully} and \texttt{harassment}, with 12 and 14 prompts misclassified in the two directions respectively; these categories share insult- and threat-related entities, so the entity block provides little discriminative signal between them. \texttt{Bully} is also occasionally predicted as \texttt{hate\_speech} (7 prompts), consistent with bullying that targets a protected group. In contrast, \texttt{adult\_content} is rarely confused with any other class.

%--------------------------------------------------------------
\section{Ablation Study}
\label{sec:ablation}

We conduct ablation experiments to isolate the contribution of the NER-augmented input and to test the design choices of the entity extraction and input construction stages. Unless stated otherwise, every variant is trained with the configuration in Table~\ref{tab:hparams} and evaluated on the same development set, with macro-F1 as the primary metric. Because differences between variants can be small relative to fine-tuning variance, each variant is trained with three random seeds and we report the mean and standard deviation.

\subsection{Effect of NER Augmentation}
We first compare the full system against two reduced inputs: (i) \emph{Prompt only}, where MARBERT is fine-tuned on the raw prompt without the \texttt{[ENTITIES]} block, and (ii) \emph{Entities only}, where the classifier sees only the serialized entity block. The first isolates the benefit of the entity cues over the contextual encoder alone; the second measures how much category information the extracted entities carry by themselves.
\subsection{Effect of Entity Label Domains}
To measure the contribution of each domain, we remove one domain at a time from the GLiNER label set (and from the serialized block), re-extract entities, and retrain. A large drop when a domain is removed indicates that its entity cues are important for the corresponding category, while a negligible change suggests redundancy with the encoder's contextual representation.

\subsection{Design Choices in Entity Extraction and Serialization}

We further vary three design choices of the pipeline, changing one factor at a time from the full system:

\begin{itemize}
  \item \textbf{Serialization format.} Our default lists all 34 labels in a fixed order with explicit \texttt{none} values. We compare it with a compact format that lists only labels with at least one detected entity. The compact format shortens the input, which matters under the 256-token limit, but removes the explicit absence signal.
  \item \textbf{GLiNER confidence threshold.} We vary the threshold in $\{0.3, 0.5, 0.7\}$ to study the trade-off between recall of entity cues (low threshold) and noise from spurious entities.
  \item \textbf{Maximum sequence length.} The maximum length $\in \{128, 256, 512\}$ to test whether truncation of the entity block affects performance.
\end{itemize}

\subsection{Discussion}
The experimental results demonstrate that the proposed pipeline achieves promising performance in fine-grained harmful content detection for Arabic user prompts. On the development set of 2,405 prompts, the system obtains a macro-F1 score of 0.886 and an accuracy of 0.897. The macro-averaged precision and recall are 0.885 and 0.888, respectively, indicating a relatively balanced classification performance across the seven categories. Since macro-F1 gives equal importance to each category, it provides a useful assessment in the presence of the class imbalance observed in the training data. Nevertheless, the overall scores conceal meaningful differences in performance between individual categories.

\subsection{Category-Specific Performance}

The system achieves its strongest performance on \texttt{adult\_content}, with an F1-score of 0.987. This result suggests that the linguistic cues associated with this category are comparatively distinguishable within the evaluated dataset. The \texttt{hate\_speech} category also achieves a high F1-score of 0.923, despite having the smallest support in the development set, with 135 instances. Similarly, \texttt{self-harm} and \texttt{FraudandDeception} obtain F1-scores of 0.902 and 0.896, respectively. These results indicate that the system can recognize several distinct types of harmful content with relatively high consistency.

In contrast, \texttt{harm-to-others} is the most challenging category, with an F1-score of 0.815, followed by \texttt{Bully} at 0.834 and \texttt{harassment} at 0.848. These differences suggest that classification difficulty is not determined solely by the number of available examples. Instead, the semantic relationship between categories and the contextual interpretation required to distinguish them appear to be important factors. In particular, categories involving threats, insults, violence, and physical injury may share similar lexical and entity-level cues, making their boundaries difficult to learn from surface information alone.

\subsection{Analysis of Classification Errors}

The confusion matrix provides further insight into the limitations of the proposed approach. The most notable confusion occurs between \texttt{harm-to-others} and \texttt{self-harm}. Specifically, 28 instances of \texttt{harm-to-others} are classified as \texttt{self-harm}, while 27 instances of \texttt{self-harm} are classified as \texttt{harm-to-others}. Both categories may contain references to physical injury, weapons, violent actions, or other forms of harm. However, distinguishing between them often requires identifying the intended target of the harmful action and interpreting the relationship between the entities and the surrounding context. Entity extraction can identify relevant expressions, but the presence of a particular entity does not necessarily establish whether the intended harm concerns the user or another person.

Another source of error is the confusion between \texttt{Bully} and \texttt{harassment}. The confusion matrix records 12 bullying instances classified as harassment and 14 harassment instances classified as bullying. This overlap is understandable because both categories can involve insults, threats, and offensive expressions. Although the entity-augmented representation explicitly identifies these cues, such information may not be sufficient to distinguish the communicative purpose, target, or situational context of an utterance. The model must therefore rely on contextual evidence beyond the presence of shared entities.

The \texttt{harm-to-others} category also receives incorrect predictions from several other categories, including \texttt{Bully}, \texttt{FraudandDeception}, and \texttt{harassment}, with 18 instances from each of these categories classified as \texttt{harm-to-others}. This pattern suggests that the category is associated with a broad range of potentially harmful expressions and may overlap with other forms of harmful behavior. In addition, seven bullying instances are classified as \texttt{hate\_speech}, which may reflect cases in which offensive behavior targets a protected group. These findings emphasize the importance of modeling semantic relationships and contextual intent when distinguishing closely related harmful-content categories.

By comparison, \texttt{adult\_content} exhibits relatively few classification errors, with only eight of its 657 development instances assigned to other categories. This comparatively clear separation suggests that some categories have more distinctive cues than others. Overall, the error analysis indicates that future improvements should focus particularly on categories whose definitions overlap semantically rather than treating all classification errors as equivalent.

\subsection{Implications of Integrating NER with MARBERT}

The proposed framework combines two complementary components: GLiNER identifies harm-related entities, while MARBERT performs contextual classification using the original prompt enriched with a structured entity representation. The entity extraction stage makes selected cues explicit, including references to violence, self-injury, financial credentials, protected groups, and offensive expressions. The classification stage then processes these cues alongside the original text rather than relying exclusively on the extracted entities.

This design provides a systematic way to incorporate entity-level information into Arabic harmful-content classification. Retaining the original prompt is particularly important because isolated entities may not reveal the intent of a message. For example, mentioning a weapon or a harmful action does not necessarily indicate a request to cause harm; the surrounding context determines how the reference should be interpreted. Combining the original prompt with the extracted entity information allows the classifier to consider both the broader linguistic context and explicit entity cues.

However, the observed results do not, by themselves, establish that entity augmentation outperforms a MARBERT-only baseline. Such a conclusion requires a direct comparison with the prompt-only configuration under the same experimental conditions. The ablation study is intended to investigate this question, as well as the effects of entity-label domains, confidence thresholds, serialization formats, and maximum sequence length. The corresponding comparisons should be reported with complete numerical results before drawing firm conclusions about the individual contribution of each design choice.

\section{Limitations and Future Directions}

Several limitations should be considered when interpreting these findings. First, the evaluation discussed here is based on the ArGuard development set. Although it provides evidence of the system's performance on the available evaluation data, further testing on the official test set and additional Arabic harmful-content datasets would be necessary to assess generalization across different sources and linguistic conditions.

Second, the performance differences between categories indicate that the system remains sensitive to semantic overlap. Future work could investigate methods that explicitly model the target and intent of harmful actions, incorporate relationships between extracted entities, or use category-specific training strategies to distinguish closely related labels. Such approaches may be particularly useful for separating \texttt{harm-to-others} from \texttt{self-harm} and distinguishing \texttt{Bully} from \texttt{harassment}.

Third, the quality of the enriched input depends on the entities detected by GLiNER. Missed entities may remove potentially useful evidence, while incorrectly extracted entities may introduce irrelevant information into the classifier's input. The fixed serialization of 34 entity labels also increases the input length, making the representation sensitive to the selected sequence-length limit. Future experiments should therefore examine entity extraction errors, the impact of alternative serialization strategies, and the balance between additional entity information and input efficiency.

Finally, because the system addresses safety-related content, classification performance should not be interpreted as a complete measure of practical safety. Errors may have different consequences depending on the category and application. Future evaluations could therefore examine category-specific error costs, calibration, robustness to dialectal variation, and performance on prompts whose harmful intent is expressed indirectly or ambiguously.

In summary, the results support the feasibility of combining generalist entity extraction with an Arabic pretrained transformer for fine-grained harmful-content classification. The system achieves strong overall development-set performance, while the error analysis identifies important challenges involving target identification, contextual intent, and overlapping category definitions. Addressing these challenges and completing the planned ablation comparisons are important steps toward establishing the effectiveness and broader applicability of the proposed framework.

\bibliography{custom}

\end{document}